\documentclass{article}

\usepackage[final]{neurips_2019}

\usepackage[utf8]{inputenc} 
\usepackage[T1]{fontenc}    
\usepackage{hyperref}       
\usepackage{url}            
\usepackage{booktabs}       
\usepackage{amsfonts}       
\usepackage{nicefrac}       
\usepackage{microtype}      

\usepackage{graphicx}
\usepackage{csquotes}
\usepackage{todonotes}
\usepackage{xspace}
\usepackage{amsmath}
\usepackage{amssymb}
\usepackage{float}
\usepackage{wrapfig}
\usepackage{lipsum}

\newcommand{\code}[1]{\texttt{#1}}
\newcommand{\cmd}[1]{\textbf{\small{\code{#1}}}}

\newcounter{lintr}

\newcommand{\twkg}{\texttt{\textsc{TextWorld KG}}\xspace}

\newcommand{\tf}{\textbf{TF-F$_1$}\xspace}
\newcommand{\fr}{\textbf{FR-F$_1$}\xspace}

\definecolor{color1}{HTML}{da6752}
\definecolor{color2}{HTML}{5573a6}
\definecolor{color3}{HTML}{6f9f6a}
\definecolor{color4}{HTML}{f3905c}
\definecolor{orange}{HTML}{ff7700}
\definecolor{cyan}{HTML}{00ffff}
\definecolor{blue}{HTML}{0000ff}

\title{Building Dynamic Knowledge Graphs from Text-based Games}

\author{Mikul\'{a}\v{s} Zelinka$^\dag$\thanks{Equal contribution, work performed when first author was an intern at Microsoft Research, Montr\'{e}al.}, Xingdi Yuan$^\ddag$\footnotemark[1], Marc-Alexandre C\^ot\'{e}$^\ddag$\footnotemark[1] \\
\textbf{Romain Laroche$^\ddag$, Adam Trischler$^\ddag$}\\
$^\dag$Charles University, Faculty of Mathematics and Physics, Czech Republic \\
$^\ddag$Microsoft Research, Montr\'{e}al \\
eric.yuan@microsoft.com
}

\begin{document}

\maketitle

\begin{abstract}
We are interested in learning how to update Knowledge Graphs (KG) from text.
In this preliminary work, we propose a novel Sequence-to-Sequence (Seq2Seq) architecture to generate elementary KG operations.
Furthermore, we introduce a new dataset for KG extraction built upon text-based game transitions (over 300k data points).
We conduct experiments and discuss the results.
\end{abstract}

\section{Introduction}

Text-based games are complex, interactive simulations in which text describes the game state and players make progress by entering text actions. 
They can be seen as sequential decision making tasks where accomplishing certain goals earns rewards (points).
Solving these games requires both Reinforcement Learning (RL) and Natural Language Processing (NLP) techniques. 

Given the complex, partially observable nature of text-based games, an explicit structured memory -- e.g., in the form of a graph -- is a useful component for game-playing agents.
In this work, we side step the game-playing aspect of the problem to focus solely on learning how to build and dynamically maintain a KG from text observations. Specifically, our proposed model learns to generate graph update operations to update an existing KG given new text information. We see this KG module as an independent block that can be leveraged by game-playing agents to improve their performance.



\textbf{Related Work:}
Numerous recent works focus on learning or using KGs in textual environments.
\cite{das18dynamickg} leverage a machine reading comprehension (MRC) mechanism to query for entities and states in short text passages and use attention to address aliased entity occurrences and to track the entity states dynamically.
\cite{xiong2018one} focus on one-shot learning of new relations from only one training instance.

For text games specifically, \cite{ammanabrolu19graph} leverage KGs to improve performance, relying on OpenIE for entity extraction and several game-specific rules for building and maintaining the KG.
\cite{sinha2019clutrr} introduce a dataset aimed at natural language understanding and generalization in reasoning about entity relations that was built using a KG-like structure.

To the best of our knowledge, all of the approaches for learning KGs are either concerned with building static KGs (rather than focusing on small, dynamic updates), or employ some domain-specific knowledge or rules to facilitate learning.
In contrast, our proposed model learns to generate general update operations for modifying a KG.

\begin{figure}[t!]
    \centering
    \includegraphics[width=1.0\textwidth]{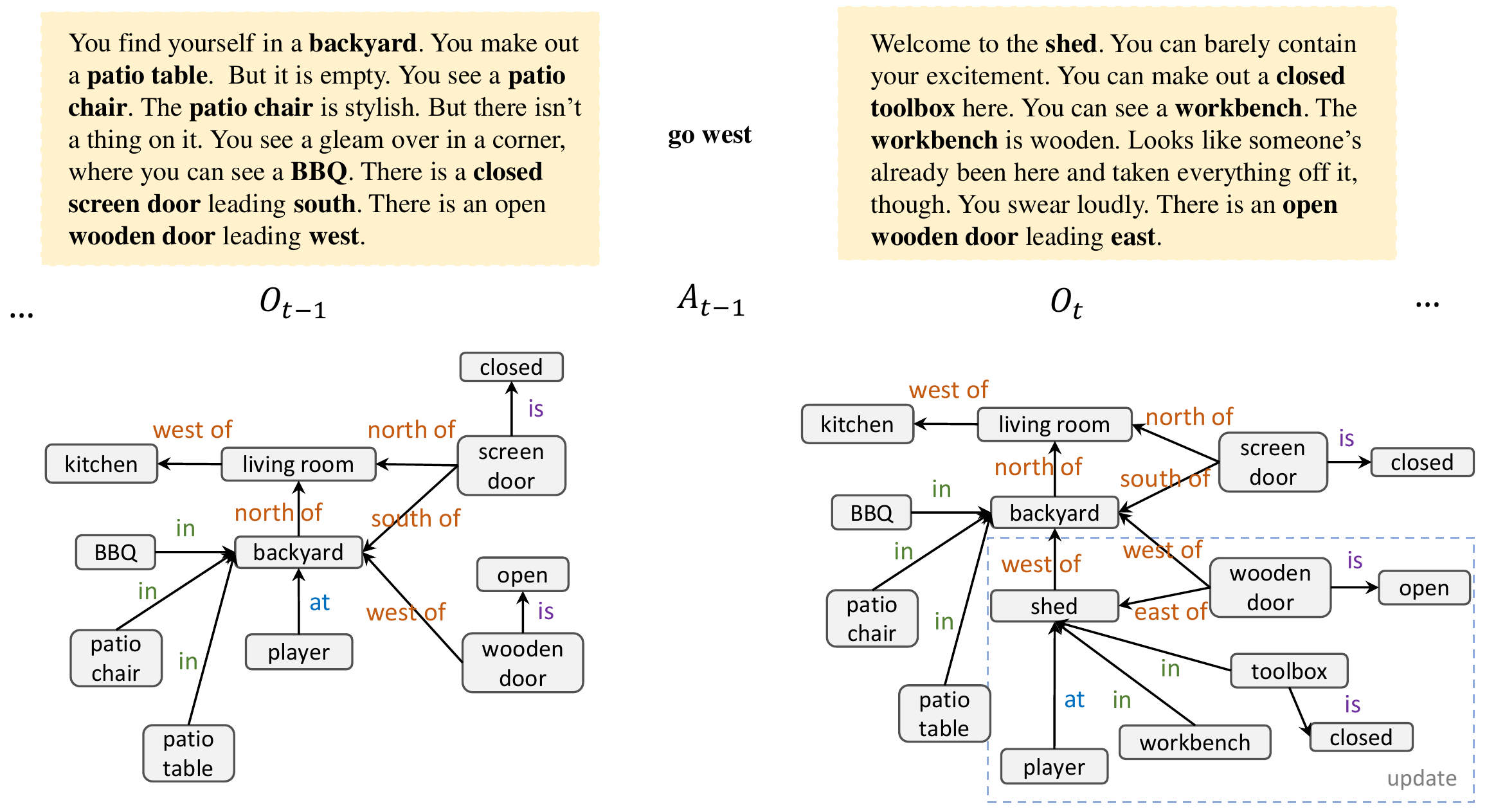}
    \caption{Illustration of an example in \twkg. By issuing an action $A_{t-1}$ at game step $t-1$, the environment returns a new observation, $O_t$. Given the KG at step $t-1$, a model is required to predict the new KG given the text observation.}
    \label{fig:kg}
\end{figure}

\section{The \twkg Dataset}
\label{section:twkg}

In this section, we introduce a new dynamic KG extraction dataset, \twkg \footnote{\twkg is publicly available at \url{https://github.com/MikulasZelinka/textworld_kg_dataset}}.
\twkg is based on a set of text-based games generated using TextWorld~\citep{cote18textworld}.\footnote{We use the games provided for the \emph{First TextWorld Problems} competition, available at\\ \url{http://aka.ms/ftwp-dataset}.}
That framework enables us to extract the underlying \emph{partial} KG for every state, i.e., the subgraph that represents the agent's partial knowledge of the world -- what it has observed so far. All games share the same overarching theme: the agent finds itself hungry in a simple modern house with the goal of gathering ingredients and cooking a meal.

To build the \twkg dataset, we collect game transitions obtained by following each game's walkthrough (provided by TextWorld). Additionally, after each step in a walkthrough, we perform 5 additional actions sampled at random from the list of admissible commands\footnote{This is the set of game actions understood by the game in a given state.} (also provided by TextWorld).
This presumably promotes robustness and generalizability of a training agent since it will encounter off-the-path transitions during game playing in the RL setting due to the absent of walkthroughs.
Therefore an agent pre-trained on such data is more likely to work well in the RL setting.
Formally, each data point in \twkg is a tuple, \{$\mathcal{G}^\text{seen}_{t-1}$, $A_{t-1}$, $O_t$, $\mathcal{G}^\text{seen}_{t}$\}, where $\mathcal{G}^\text{seen}_{t-1}$ is a partial KG (in the format of Resource Description Framework (RDF) triples) representing the information an agent has seen during all previous game steps up to $t-1$. After the agent issues an action $A_{t-1}$ (a string of words), the game engine returns a new text observation $O_t$ describing its effects on the world.
The task is to predict the updated partial KG $\mathcal{G}^\text{seen}_{t}$ given the above information.
Note that the new observation might not contain new information, since some actions do not change the game state (e.g. \cmd{look} in the same room twice).
Table~\ref{tab:stats} shows some statistics about \twkg and Figure~\ref{fig:kg} illustrates an example data point.

An important challenge posed by TextWorld is generalization. 
In each individual game instance, the interactable objects and their locations change along with the layout of the environment. 
Similarly, object names can be composed of multiple adjectives and a noun (e.g., red hot chili pepper), and at test time, players may encounter object names never seen during training. \twkg inherits both of these challenging features.

\begin{table}[h!]
    \centering
    \scriptsize
    \begin{tabular}{c|c|c||c|c||c|c|c}
        \toprule
        \#Train & \#Valid & \#Test & Avg. Obs. & Avg. \#Operations & \#Vertices & \#Edges & Avg. \#Connections \\
        \midrule
        267,031 & 13,442 & 41,865 & 29.3 tokens& 3.1 & 99 & 10 & 43.1 \\
        \bottomrule
    \end{tabular}
    \vspace{0.6em}
    \caption{Statistics of \twkg. \textbf{Avg. Obs.} is the average number of tokens an observation has. \textbf{Avg. \#Operations} is the average number of update operations to generate per time step. \textbf{\#Vertices} and \textbf{\#Edges} correspond to the number of unique entities and relation types. \textbf{Avg. \#Connections} is the average number of connections a graph has.}
    \label{tab:stats}
\end{table}

\section{Learning to Update a KG}
\label{section:learn_update_kg}

\subsection{KG Definition}
\label{section:kg_definition}

In a text-based game, at any given game step $t$, the game state $s_t$ can be represented as a graph $\mathcal{G}^\text{full}_t = (\mathcal{V}_t, \mathcal{E}_t)$. 
In our setting, vertices $\mathcal{V}_t$ represent entities (including objects, the player, and locations) and their states (e.g., closed, fried, sliced). 
Vertices are connected by edges $\mathcal{E}$, which represent a set of relations between entities (e.g. \cmd{north\_of}, \cmd{in}, \cmd{is}).

Since games are partially observable, at every step an agent only observes part of the full game state (e.g., the agent cannot know facts in a room it has not visited). 
Thus, an agent must build its belief about the world, $\mathcal{G}^\text{belief}_t$, from its observations. Ideally, the belief graph should match the ground truth graph, $\mathcal{G}^\text{seen}_t$, which is a subgraph of $\mathcal{G}^\text{full}_t$ representing what has been seen so far in the game.

TextWorld games are deterministic, so by progressively exploring and observing, an agent should discover more knowledge to push into its belief graph. Eventually, this ought to converge to a graph that accurately represents the entire game state.

\subsection{Updating a KG}
\label{section:how}

Instead of generating the entire belief graph at every game step, we generate a set of update operations $\Delta g_t$ such that $\mathcal{G}^\text{belief}_t = \text{Update}(\mathcal{G}^\text{belief}_{t-1}, \Delta g_t)$, 
where $\text{Update}$ is an oracle function that applies $\Delta g_t$. 
In our case, each update operation in $\Delta g_t$ is represented as a text command. 
We define the following two types of update operation:
\begin{itemize}
    \item \cmd{add(node1, node2, relation)}: add a directed edge, named \cmd{relation}, between \cmd{node1} and \cmd{node2}; if any of these nodes does not exist, add that node first.
    \item \cmd{delete(node1, node2, relation)}: delete a directed edge, named \cmd{relation}, between \cmd{node1} and \cmd{node2}; if any of the nodes or the edge does not exist, ignore this operation. 
\end{itemize}

Given a new observation string $O_t$ and an agent's current belief $\mathcal{G}^\text{belief}_{t-1}$, the agent is required to generate $k\geq 0$ operations as defined above to merge newly observed information into its belief graph.
For the example shown in Figure~\ref{fig:kg}, the generated operations are listed in Table~\ref{tab:example_cmds}.

\begin{wraptable}{r}{0.5\textwidth}
    \centering
    \begin{tabular}{l}
        \toprule
         \cmd{add (player, shed, at)}\\
         \cmd{add (shed, backyard, west\_of)}\\
         \cmd{add (wooden door, shed, east\_of)}\\
         \cmd{add (toolbox, shed, in)}\\
         \cmd{add (toolbox, closed, is)}\\
         \cmd{add (workbench, shed, in)}\\
         \cmd{delete (player, backyard, at)}\\
        \bottomrule
    \end{tabular}
    \caption{Update operations corresponding to the transition shown in Figure~\ref{fig:kg}.}
    \label{tab:example_cmds}
\end{wraptable}

We formulate the update generation task as a Seq2Seq problem.
Specifically, we adopt the decoding strategy from \cite{yuan18kpgen}, where given an observation sequence $O_t$ and a belief graph $\mathcal{G}^\text{belief}_{t-1}$, the agent generates a sequence of tokens consisting of multiple graph update operations separated by a delimiter token.

As pointed out by \citet{meng19order}, the order of ground-truth tokens and sequences (in our case, graph update operations) matters in Seq2Seq language generation. We therefore define a set of rules (e.g., always \cmd{add} before \cmd{delete}) to order ground-truth operations for teacher forcing during training.

\subsection{Model Architecture}
\label{section:model}

\begin{figure}[h!]
    \centering
    \includegraphics[width=0.8\textwidth]{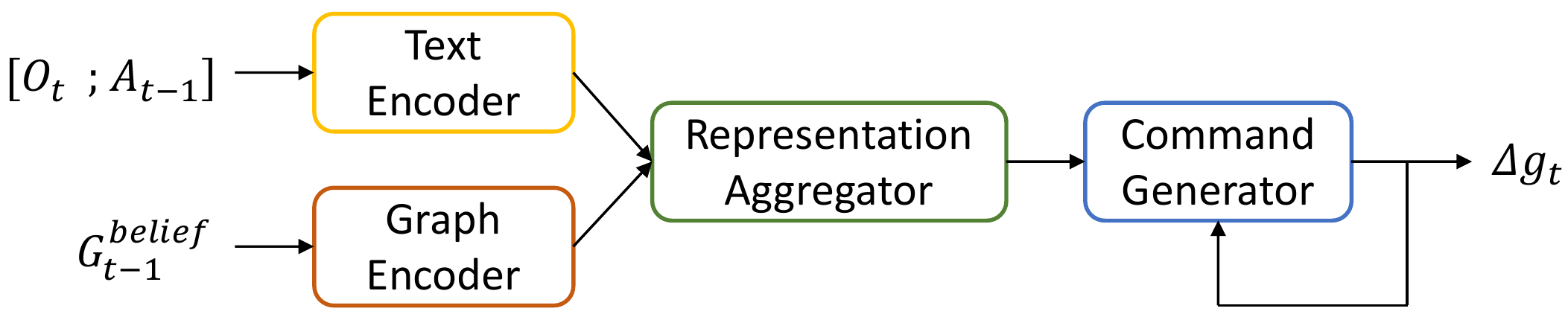}
    \caption{Graph update operation generation model.}
    \label{fig:agent}
\end{figure}

We use a transformer-based Seq2Seq model \citep{vaswani17allyouneed} to generate update operations. 
As shown in Figure~\ref{fig:agent} in Appendix~\ref{appendix:model}, the model consists of the following components: 
\begin{enumerate}
    \item A \textbf{text encoder}, which reads text inputs (the concatenation of observation $O_t$ and the action at the previous game step, $A_{t-1}$), and generates hidden representations.
    \item A \textbf{graph encoder}, which encodes the previous belief $\mathcal{G}^\text{belief}_{t-1}$ into hidden representations.
    \item An attention-based \textbf{representation aggregator}, which combines the two above representations. 
    \item A \textbf{command generator}, which takes aggregated representations and generates update operations token by token.
\end{enumerate}


For space considerations, we elaborate our model components in Appendix~\ref{appendix:model}.
Following common practice in natural language generation (NLG), we train our operation generation model via teacher forcing.
Specifically, during training, a right-shifted ground truth target sequence is provided as input to the decoder and the model is trained with the negative log-likelihood (NLL) loss.
During test, the model starts generating from a start-of-sentence token and uses the previously generated token as input to the next step. The model terminates after generating an end-of-sequence token.

\section{Experiments and Discussion}
\label{section:exp}

\begin{wraptable}{r}{0.5\textwidth}
    \centering
    \small
    \begin{tabular}{l|c|c}
        \toprule
        Model & \tf & \fr  \\
        \midrule
        \text{Transformer} & 0.832 & 0.434\\
        \midrule
        \text{\:\:\:\:\:\:\:\:+ GCN} & 0.928 & 0.664 \\
        \midrule
        \text{\:\:\:\:\:\:\:\:+ R-GCN} & \textbf{0.965} & 0.645 \\
        \midrule
        \text{\:\:\:\:\:\:\:\:\:\:\:\:\:\:\:\:+ R-Emb} & 0.962 & \textbf{0.697} \\
        \bottomrule
    \end{tabular}
    \caption{Test performance.}
    \label{tab:exp_results}
\end{wraptable}

In this preliminary study we test 4 graph encoder variants of the proposed model.
First, as a baseline, we disable the graph encoder, which renders the model a standard Seq2Seq transformer.
Second, we utilize a Graph Convolutional Network (GCN) \citep{kipf16gcn} as the graph encoder. The GCN does not consider multiple relations.\footnote{For models that do not consider relational information, we use single relation KGs as ground-truth during evaluation.}
Third, we enable conditioning on multiple relations by using a Relational Graph Convolutional Network (R-GCN) \citep{schlichtkrull2018rgcn}.
Although R-GCN takes into account multiple relations, it does not consider information in relation labels.
In our task, this information is important (e.g., \cmd{east\_of} and \cmd{west\_of} are symmetric relations). 
Therefore, we finally learn a vector representation for each relation that is conditioned on the label's text embeddings.
The resulting relation representation is used as an extra input to the R-GCN layer. Table~\ref{tab:exp_results} shows the test results for all models.

During training, a model takes \{$\mathcal{G}^\text{seen}_{t-1}$, $A_{t-1}$, $O_t$\} as input, where $\mathcal{G}^\text{seen}_{t-1}$ is the input graph, and $A_{t-1}$ and $O_t$ are the text action issued at the previous game step and the resulting text observation, respectively. The model outputs a sequence describing an update operation to the graph and the resulting $\mathcal{G}^\text{belief}_{t}$.
During evaluation: 
\begin{itemize}
    \item \textbf{Teacher-force (TF) \textbf{F$_1$}}: we use the ground-truth $\mathcal{G}^\text{seen}_{t-1}$ as input graph and compute the $\text{F}_1$ score between the model's generated graph update \textit{commands} and the ground-truth commands. Note the $\text{F}_1$ score is computed on command level (i.e., if any token in a command is incorrect, this command is treated as incorrect).
    \item \textbf{Free-run (FR) \textbf{F$_1$}}: we initialize the belief graph at the beginning of each game with an empty graph. For each game step, we use $\mathcal{G}^\text{belief}_{t-1}$ (the graph generated by the model) as input. At the end of each game, we compute $\text{F}_1$ score between the final belief graph $\mathcal{G}^\text{belief}_{T}$ and ground truth $\mathcal{G}^\text{seen}_{T}$, graphs are represented as \textit{RDF triples}.
\end{itemize}


In general, although all model variants show good performance on \tf, they perform worse on \fr. This is not surprising since errors accumulate in the latter setting.
Models using R-GCN outperform those using GCN by a noticeable margin, which suggests relational information is essential in the proposed tasks.
Interestingly, while the two R-GCN models perform similarly on \tf, the variant with relational embedding (considering information in relation labels) significantly outperforms the other on \fr.

To better understand the behavior of our proposed models on \twkg, we conduct an error analysis, which we show in Appendix~\ref{appendix:error_analysis}.

The next step for this project is to leverage the KG update module while playing text-based games. We believe that maintaining such a graph could help an RL agent (1) to avoid re-discovering known facts about the world and (2) to discover new world knowledge efficiently. We are also interested in finding ways of transferring learned graphs from one game to another to improve agents' ability to generalize.

\bibliographystyle{apalike}
\bibliography{biblio}

\appendix

\section{Model Architecture}
\label{appendix:model}

\paragraph{Text Encoder:}
An observation string $O_t$ is provided in response to the text action $A_{t-1}$ an agent issued at previous game step.
We concatenate the two text information together $[A_{t-1} ; O_t]$ as the input to the text encoder, in which $[\cdot;\cdot]$ indicates vector/string concatenation. 
The text encoder consists an embedding layer and a stack of transformer blocks.
The text encoder results a sequence of hidden vectors $h^O \in \mathbb{R}^{L^O \times H}$, where $L^o$ is length of the concatenated string, $H$ is hidden size.

\paragraph{Graph Encoder:}
At the same time, the graph encoder takes the agent's belief KG $\mathcal{G}^\text{belief}_{t-1}$ (in which stores the agent's observations a all previous game steps) as input. 
We adopt different off-the-shelf graph neural networks (GNN) as the graph encoder (we will describe more details in experiment section).
After several layers of propagation, graph representations $h^\mathcal{G} \in \mathbb{R}^{N \times H}$ are generated, where $N$ is number of nodes in the KG.

\paragraph{Representation Aggregator:}
We use an attention based layer to aggregate text and graph representations \citep{bahdanau2014attention,seo2016bidaf}. 
Specifically, we use weighted sum of text representations to represent graph information, which results $h^{\mathcal{G}O} \in \mathbb{R}^{N \times H}$; similarly, we use weighted sum of graph representations to represent text information, resulting $h^{O\mathcal{G}} \in \mathbb{R}^{L^O \times H}$.

\paragraph{Command Generator}
Finally, both $h^{O\mathcal{G}}$ and $h^{\mathcal{G}O}$ are used to condition text generation. 
Input tokens are first converted into embeddings, then they are fed into a stack of decoder transformer blocks which result a probability distribution over the vocabulary.
To prevent model from utilizing future information, we follow \cite{vaswani17allyouneed} to use a masked multi-head self attention layer in the beginning of each block.
For being able to both generate a word from vocabulary, and point a word from the source text $O_t$, we adopt the pointer-softmax mechanism \citep{gulcehre16pointer}.

\section{Error Analysis}
\label{appendix:error_analysis}

\begin{figure}[h!]
    \centering
    \includegraphics[width=1.0\textwidth]{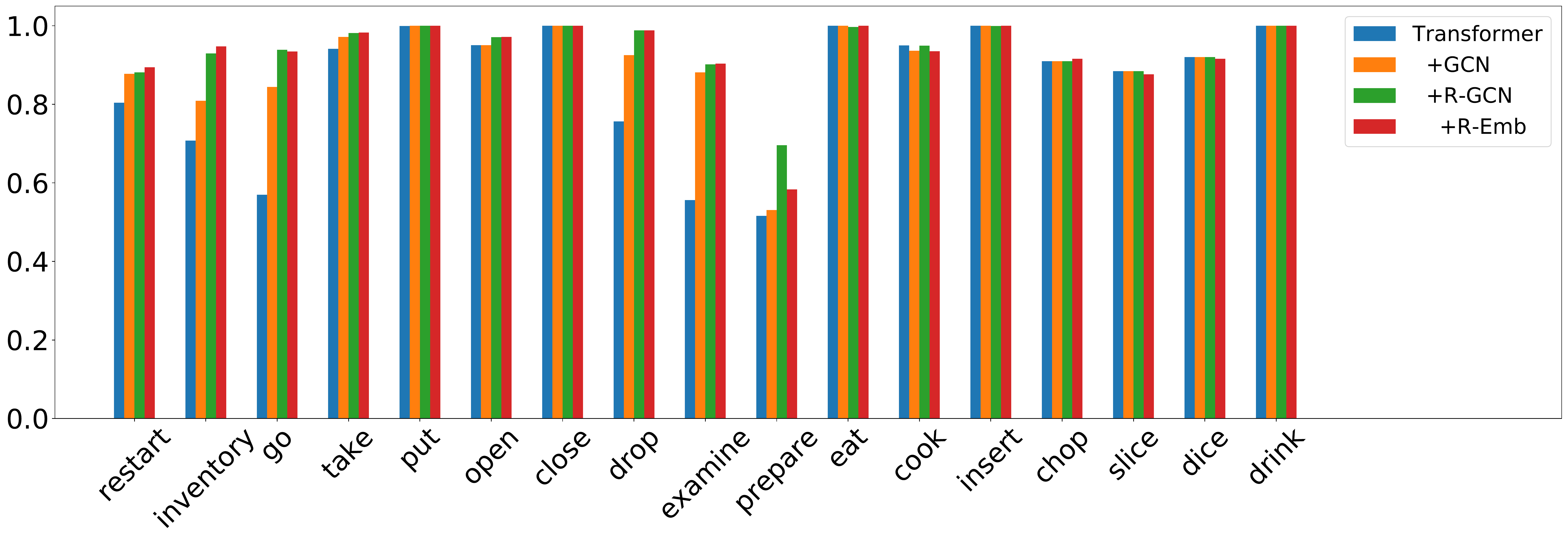}
    \caption{Average \tf scores grouped by verbs.}
    \label{fig:error}
\end{figure}

In Figure~\ref{fig:error}, we report average \tf scores grouped by verbs found in input actions $A_{t-1}$.
We can observe that vanilla transformer model performs poorly on \cmd{go} actions, which aligns with the fact that after issuing a \cmd{go} action, the resulting observation text $O_t$ does not contain any information of the agent's previous location. On the other hand, the other models benefit from their belief graph to retrieve that \emph{single} information.

Also, we notice that all models perform relatively poorly on \cmd{prepare} actions. 
This also makes sense since in TextWorld games, the action of preparing a meal consumes multiple food ingredients at once in order to produce a meal object.
The resulting observation text $O_t$ following a \cmd{prepare} action only contains information about the newly produced meal and does not mention what food ingredients have been consumed.
Even though the information about the recipe (i.e. ingredients needed) is part of the KG graph, a model has to learn to retrieve \emph{multiple} information at once from its belief graph $\mathcal{G}^\text{belief}_{t-1}$ to figure out what ingredients will be consumed.

\end{document}